\documentclass[10pt, journal, twoside, twocolumn]{IEEEtran}

\usepackage{algorithm}
\usepackage{algpseudocode}
\usepackage{array}
\usepackage[caption=false,font=normalsize,labelfont=sf,textfont=sf]{subfig}
\usepackage{textcomp}
\usepackage{stfloats}
\usepackage{url}
\usepackage{verbatim}
\usepackage{graphicx}
\usepackage{cite}
\usepackage{picinpar}
\usepackage{amsmath,amsfonts}
\usepackage{url}
\usepackage{flushend}
\usepackage[latin1]{inputenc}
\usepackage{colortbl}
\usepackage{soul}
\usepackage{multirow}
\usepackage{pifont}
\usepackage{color}
\usepackage{alltt}

\usepackage{enumerate}
\usepackage{siunitx}
\usepackage{breakurl}
\usepackage{epstopdf}
\usepackage{pbox}

\usepackage{array}
\usepackage[caption=false,font=normalsize,labelfont=sf,textfont=sf]{subfig}
\usepackage{textcomp}
\usepackage{stfloats}
\usepackage{verbatim}
\usepackage{xcolor}
\usepackage{booktabs}
\usepackage{gensymb}
\usepackage[hidelinks]{hyperref}
\usepackage{threeparttable}

\usepackage{amssymb} 
\usepackage{tikz}
\usepackage{listings}
\usepackage{authblk}

\newcommand{\figref}[2]{Fig.~\hyperref[#1]{\ref*{#1}(#2)}}

\newcolumntype{P}[1]{>{\centering\arraybackslash}p{#1}}

\newcolumntype{C}[1]{>{\centering\arraybackslash}p{#1}}

\def\BibTeX{{\rm B\kern-.05em{\sc i\kern-.025em b}\kern-.08em
    T\kern-.1667em\lower.7ex\hbox{E}\kern-.125emX}}
\usepackage{balance}

\begin{document}
\title{DART: Dual-level Autonomous Robotic Topology for Efficient Exploration in Unknown Environments}
\author{Qiming Wang, Yulong Gao\textit{, Senior Member, IEEE}, Yang Wang, Xiongwei Zhao\textit{, Student Member, IEEE}, \\ 
	\vspace{-2ex}
	Yijiao Sun and Xiangyan Kong
	\thanks{This work was supported by the National Natural Science Foundation of China under Grant 62171163.}
	\thanks{Qiming Wang, Yulong Gao and Xiangyan Kong are with the School of Electronic and Information Engineering, Harbin Institute of Technology, Harbin 150001, China. (e-mail: \href{21b905005@stu.hit.edu.cn}{21b905005@stu.hit.edu.cn}; \href{ylgao@hit.edu.cn}{ylgao@hit.edu.cn}; \href{22b905071@stu.hit.edu.cn}{22b905071@stu.hit.edu.cn})}
	\thanks{Yang Wang, Xiongwei Zhao and Yijiao Sun are with the School of Electronic and Information Engineering, Harbin Institute of Technology (Shenzhen), Shenzhen 518055, China. (e-mail: {\href{yangw@hit.edu.cn}{yangw@hit.edu.cn}; \href{xwzhao@stu.hit.edu.cn}{xwzhao@stu.hit.edu.cn};  \href{sunyijiao@stu.hit.edu.cn}{sunyijiao@stu.hit.edu.cn}})}
}


\maketitle

\begin{abstract}
	Conventional algorithms in autonomous exploration face challenges due to their inability to accurately and efficiently identify the spatial distribution of convex regions in the real-time map. These methods often prioritize navigation toward the nearest or information-rich frontiers---the boundaries between known and unknown areas---resulting in incomplete convex region exploration and requiring excessive backtracking to revisit these missed areas. To address these limitations, this paper introduces an innovative dual-level topological analysis approach. First, we introduce a Low-level Topological Graph (LTG), generated through uniform sampling of the original map data, which captures essential geometric and connectivity details. Next, the LTG is transformed into a High-level Topological Graph (HTG), representing the spatial layout and exploration completeness of convex regions, prioritizing the exploration of convex regions that are not fully explored and minimizing unnecessary backtracking. Finally, an novel Local Artificial Potential Field (LAPF) method is employed for motion control, replacing conventional path planning and boosting overall efficiency. Experimental results highlight the effectiveness of our approach. Simulation tests reveal that our framework significantly reduces exploration time and travel distance, outperforming existing methods in both speed and efficiency. Ablation studies confirm the critical role of each framework component. Real-world tests demonstrate the robustness of our method in environments with poor mapping quality, surpassing other approaches in adaptability to mapping inaccuracies and inaccessible areas.
\end{abstract}

\section{Introduction}

\IEEEPARstart{A}{utonomous} exploration is the process of controlling a sensor-equipped platform to actively navigate in an unknown environment, with the goal of systematically exploring and mapping all unknown regions \cite{SLAM_1}, while minimizing the completion time and travel cost of the exploration task. This technology is widely applied in search and rescue \cite{ref2}, scene reconstruction \cite{ref3}, planetary exploration\cite{ref4} and industrial inspection\cite{ref5}, significantly improving safety in hazardous environments and enhancing the efficiency of operational tasks. The main challenge in autonomous exploration lies in analyzing the map generated by the SLAM algorithm to determine the optimal navigation target point immediately after each map update. However, conventional exploration methods still exhibit significant shortcomings, including low exploration efficiency and inadaptability to complex environments. These issues can be specifically identified as the following three problems: 
\begin{enumerate}[1)]
	\item \textbf{Inefficient Frontier Extraction and Unverified Accessibility}: Current autonomous exploration methods involve comprehensive scanning of the entire map to identify frontiers \cite{frontier1,frontier1m,frontier2,frontier3,frontier6,frontier8}, resulting in a heavy computation burden. Additionally, these methods often neglect to assess the accessibility of these frontiers, which can lead to exploration failures, particularly when robots are navigated through routes that are impassable.
	\item \textbf{Greedy Exploration Strategy}: Using a greedy utility function that weighs potential information gain against travel costs, conventional exploration algorithms often lead robots to regions with high information gain before completing the exploration of the current region \cite{frontier4, frontier9, sample3, sample6, DRL6, DRL8}. This approach results in backtracking to previously missed regions and reduces global exploration efficiency.
	\item \textbf{Low Responsiveness to Changing Navigation Targets}: Contemporary algorithms primarily focus on enhancing decision-making for candidate targets \cite{frontier4, sample4, sample11, DRL8} and optimizing navigation trajectories \cite{frontier7, DRL1, DRL2}. However, most still rely on traditional local path planning algorithms, such as DWA \cite{dwa} and TEB \cite{teb}, for motion control. This reliance often results in slower response times to frequently changing navigation targets, which are typical in exploration tasks as maps are continually updated.
\end{enumerate}

Therefore, there is a critical need for advanced algorithms designed to address these specific problems. 

\begin{figure}[t]
	\centering
	\vspace{6pt}
	\includegraphics[width=\columnwidth]{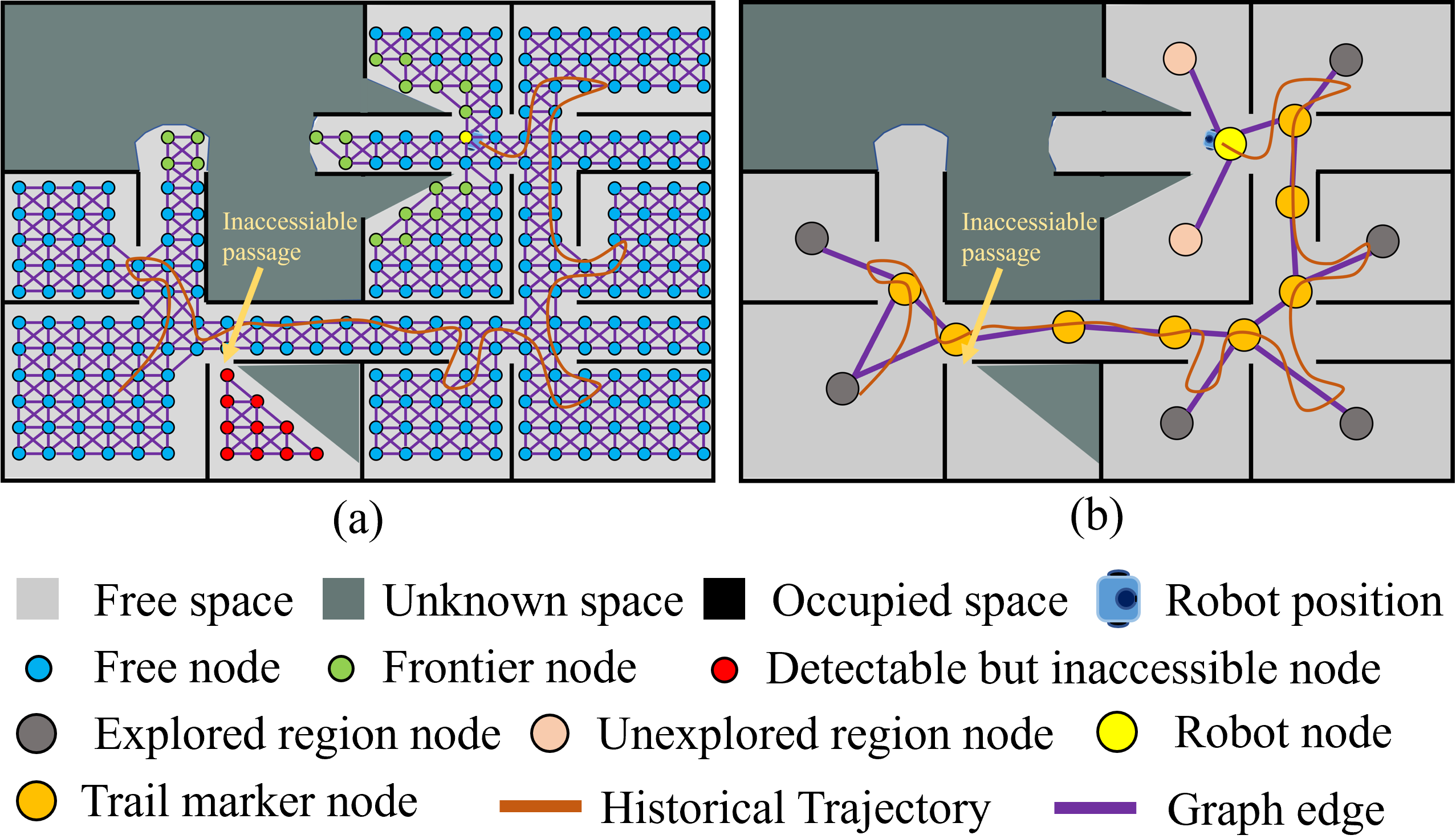}
	\caption{Illustration of the dual-level topological exploration Graph. (a) The LTG, generated by uniformly sampling the original map, serves as the foundation for subsequent exploration algorithms. (b) The HTG, derived from processing the LTG, shows not only the distribution and connectivity of the convex polygon regions but also whether each region has been fully explored.}
	\label{fig_1}
\end{figure}

Recognizing these challenges, our framework introduces an innovative dual-level topological analysis approach, as shown in Fig. \ref{fig_1}. Firstly, we transform the original map into a Low-level Topological Graph (LTG) using a uniform sampling method, which serves as the foundation of the entire framework. This transformation succinctly captures the geometric and connectivity information of the spatial layout within the known region of the map, as shown in \figref{fig_1}{a}. Secondly, the High-level Topological Graph (HTG) is constructed from the LTG using an innovative graph-based erosion method inspired by binary erosion techniques in computer graphics. This provides a global topological structure of the convex polygonal regions within the known area of the map, optimizing exploration strategies by prioritizing the exploration of incomplete convex polygonal regions, as shown in \figref{fig_1}{b}. Thirdly, we propose an innovative Local Artificial Potential Field (LAPF) approach for motion control, which generates motion control commands quickly based on the node path generated from the LTG once the navigation target is determined. This approach addresses the challenge of the robot needing to rapidly respond to frequent target point changes as the map is updated, improving exploration efficiency. Finally, our method not only enhances the strategic prioritization of target exploration areas but also reduces computational complexity, facilitating rapid and efficient exploration in environments with complex spatial distributions and structural configurations.

In summary, this paper presents four main contributions; the initial three sequentially tackle the problems identified earlier.

\begin{enumerate}[1)]
	\item \textbf{LTG for Rapid Frontier Extraction and Accessibility Analysis}: We proposed the Low-level Topological Graph (LTG), a novel approach derived from uniform sampling of the original map. LTG prioritizes peripheral nodes to facilitate rapid frontier detection and clustering. Additionally, it incorporates connectivity checks when establishing edges between adjacent nodes, effectively eliminating impassable pathways and enhancing navigational safety.
	\item \textbf{HTG for non-greedy exploration strategy}: Utilizing the High-level Topological Graph (HTG), our approach segments the map into convex polygonal region. This method ensures that the robot completes the exploration of the current region before advancing to the next, making navigation decisions independent of the information gain of the frontiers.
	\item \textbf{LAPF Motion Control}: Leveraging the structural data from the LTG, Local Artificial Potential Field (LAPF) dynamically optimizes movement decisions, enhancing navigation efficiency and improving the robot's response speed to frequent target point changes.
	\item \textbf{Empirical Validation and Adaptability}: Extensive simulations and real-world tests have confirmed our method's enhancements in exploration speed and computational efficiency, demonstrating significant reductions in time and distance while highlighting its robustness and adaptability to complex environments.
\end{enumerate}

This article is structured as follows: Section II surveys existing research in robotic autonomous exploration. Section III elucidates the problem description. Section IV elaborates on our novel framework. Section V presents a comprehensive analysis of experimental results. The article concludes with a summary in Section VI.

\section{Related Work}

As the widespread application of autonomous exploration continues, extensive research has been conducted to address the challenge of minimizing the time and path length costs of the overall exploration task. Existing autonomous exploration algorithms can be classified into three categories: frontier-based methods \cite{frontier1,frontier1m, frontier2,frontier3,frontier4,frontier5,frontier6,frontier17,frontier7,frontier8,frontier9,frontier10,frontier11,frontier12,frontier13,frontier14,frontier15,frontier16}, sample-based methods \cite{sample1,sample2,sample3,sample4,sample5,sample6,sample7,sample8,sample9,sample10, sample11, sample12}, and deep reinforcement learning-based (DRL-based) methods \cite{DRL1,DRL2,DRL3,DRL4,DRL5,DRL6,DRL7,DRL8,DRL9,DRL10,DRL11,DRL12,DRL13}.

The frontier-based methods involve extracting and clustering the frontier---the boundary between known and unknown areas---of the real-time map, designing a utility function to calculate the value of each cluster, and selecting the cluster with the highest utility as the navigation target. In the initial work on frontier-based methods \cite{frontier1,frontier1m}, utility functions were designed simply, leading robots to select the frontier cluster with the minimal distance cost as their target, frequently resulting in inefficiencies in exploration tasks. To address this issue, many researchers have refined the methodology. In \cite{frontier2,frontier3,frontier4,frontier5,frontier6,frontier17}, methods have been designed for the rapid extraction and clustering of frontiers, along with a specific utility function that inclines robots to select a frontier cluster with high potential information gain as the navigation target, achieving improved exploration efficiency. In addition to frontier extraction and utility function optimization, trajectory optimization has also become a research hotspot. \cite{frontier7,frontier8,frontier9,frontier10,frontier11,frontier12} propose methods to optimize the trajectory after selecting the target point, thereby enabling the path to explore more potential regions or reach the target point faster. Furthermore, \cite{frontier13,frontier14,frontier15,frontier16} focus on segmenting maps of structured scenes, thereby predicting the potential information of feature regions to optimize the exploration sequence of frontiers in different characteristic areas. However, although these methods achieve a high exploration rate in the early stages, the neglect of frontiers with smaller information gains necessitates extensive backtracking in later stages to revisit these previously missed areas, resulting in increased total exploration time and travel distance.

The sample-based methods are focused on determining the target points by using the Rapidly exploring Random Tree (RRT) \cite{RRT} algorithm to build a node graph on a real-time map and analyzing the exploration efficiency of each particular node. Since sample-based methods generally only need to analyze the vicinity of the sampled nodes rather than the entire map, this method is computationally fast and takes up little storage space. In early work, \cite{sample1} introduced the next-best-view planner (NBVP), incorporating the method of the next-best-view (NBV) \cite{sample2} and RRT to enhance autonomous exploration. Based on this method, extensive research has been conducted to refine it \cite{sample3,sample4,sample5,sample6,sample7,sample8}; for example, \cite{sample3} uses RRT to analyze connectivity between nodes after sampling on a occupancy grid map to determine path feasibility and generate frontier; \cite{sample4} treats nodes as viewpoints to determine the amount of potential information of node paths in 3D environments, and uses the method of receding horizon fashion to ensure real-time path generation. In addition, the integration of semantic information with the RRT technique has emerged as a significant area of research \cite{sample9,sample10, sample11, sample12}. These approaches involve storing semantic data from each viewpoint in sampling nodes, enabling detailed analysis of the exploration area's potential information. Nevertheless, due to the randomness inherent in sampling-based methods, these approaches can result in missed detections of critical information areas. Similarly to frontier-based methods, they also tend to prioritize regions with high potential information, leading to extensive backtracking in later stages of exploration.

The DRL-based methods have become a research hotspot in recent years \cite{RL1,RL2}. These methods can be categorized into two main types: hybrid methods that integrate traditional methodologies \cite{DRL1,DRL2,DRL3,DRL4}, and purely model-based methods that rely entirely on neural network models to guide exploration decisions \cite{DRL5,DRL6,DRL7,DRL8,DRL9,DRL10,DRL11,DRL12,DRL13}. Hybrid approaches depend on conventional exploration strategies for selecting target points, subsequently employing deep reinforcement learning to generate navigation trajectories or motion control commands. For example, \cite{DRL1,DRL2} combine frontier-based methods with Voronoi diagrams to allocate navigation target points to each robot, and then employ deep reinforcement learning models to generate motion commands. Purely model-based methods are categorized into two-stage \cite{DRL5,DRL6,DRL7,DRL8} and end-to-end approaches \cite{DRL9,DRL10,DRL11,DRL12,DRL13,DRL14}. \cite{DRL8} proposes a two-stage method, utilizing a dual-layer model where the top layer decides on target points and the bottom layer handles navigation. \cite{DRL14} adopts an end-to-end approach, configuring a model that guides robots towards navigable and information-rich target points. However, DRL-based methods still fail to address the inherent greediness in autonomous exploration tasks, where robots tend to explore areas with high potential information while neglecting less informative yet crucial regions, ultimately leading to subsequent backtracking.

Existing autonomous exploration techniques often exhibit greedy behaviors that result in inefficiencies, including unnecessary backtracking and redundant exploration. In response, our approach introduces a dual-level topological graph that analyzes the spatial distribution of convex polygon regions on the map to improve exploration strategy by systematically exploring all unexplored regions in spatial sequence.

\section{Problem Description}
This research addresses the challenge of exploring a bounded two-dimensional area \( V \subset \mathbb{R}^2 \), while simultaneously generating an accurate occupancy grid map. The occupancy grid \( M \) subdivides \( V \) into square grids \( g \in M \), and each grid is classified into three states based on integrated data from robotic sensory observations and computations from a SLAM algorithm. The states for each grid are defined as follows:

\begin{itemize}
	\item \( g_{\text{free}} \) if the probability of being unoccupied is high,
	\item \( g_{\text{occ}} \) if the probability of being occupied is high,
	\item \( g_{\text{unk}} \) if neither state can be confidently determined.
\end{itemize}

The status of each grid \( g \) is dynamically updated through a synthesis of sensor data and the inferential capabilities of the SLAM algorithm, which assesses the likelihood of each state based on the cumulative data gathered during exploration.

Solving the exploration problem effectively entails achieving comprehensive coverage of \( V \), defined mathematically by:
\begin{equation}
	\text{Coverage}(V) = \frac{\sum_{g \in M} I_{\text{known}}(g)}{|M|} \times 100\%
\end{equation}
where \( I_{\text{known}}(g) \) is an indicator function that returns 1 if the state of the grid \( g \) is known (either \( g_{\text{free}} \) or \( g_{\text{occ}} \)), and 0 if it remains \( g_{\text{unk}} \). The goal is to maximize the coverage metric, ideally reaching close to 100\% for a fully explored and accurately mapped environment, while minimizing the time required to achieve this coverage.

\section{Methodology}

\begin{figure*}[t] 
	\centering
	\includegraphics[width=\textwidth]{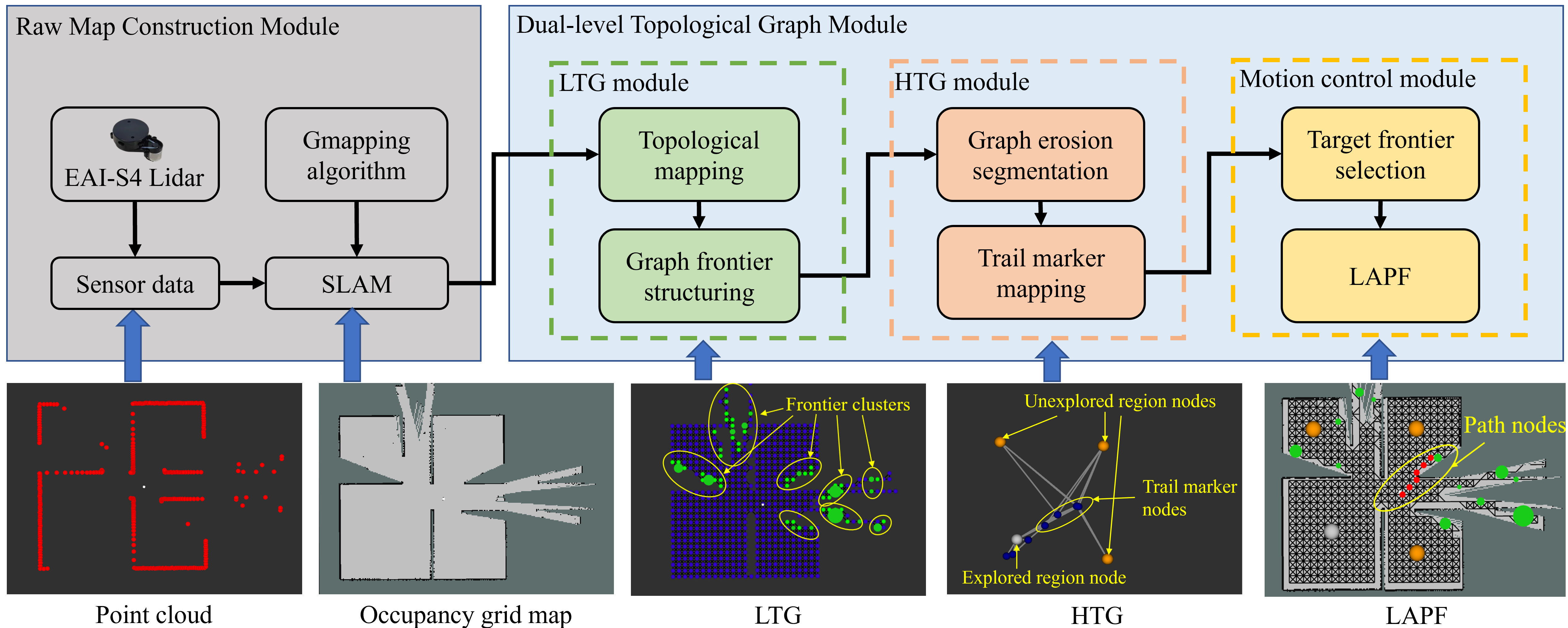} 
	\caption{The dual-level topological graph framework comprises three main components. The first component, the LTG module, is constructed through uniform sampling of the occupancy grid map to extract frontier information. The second component, the HTG module, integrates trail marker nodes with region nodes derived by eroding the LTG, facilitating structured exploration planning. The final component, the motion control module, employs a LAPF method to directly command the robot's movements, optimizing navigation based on real-time data and graph information.}
	\label{fig_2}
\end{figure*}

This section presents the dual-level topological graph framework as shown in Fig. \ref{fig_2}. The robot first processes sensor data with a SLAM algorithm to generate an occupancy grid map, which is uniformly sampled to form the LTG for frontier node detection and clustering. Then, erosion-based segmentation is used to extract convex region nodes, which are combined with trail marker nodes to generate the HTG. The exploration strategy initially prioritizes unexplored convex regions within the HTG. Once all convex regions have been explored, it shifts focus to clustered frontiers within the LTG. Finally, the LAPF method generates motion control commands by analyzing path nodes and environmental data to optimize navigation.

\subsection{The LTG Construction and Node-Based Frontier Detection}
The process of the LTG module (the green block in Fig. \ref{fig_2}) is detailed in Algorithm~\ref{alg:low_level_topo}. This subsection comprises two main modules: the Topological Mapping module (Algorithm~\ref{alg:low_level_topo}, lines 6-14) and the Graph Frontier Structuring module (Algorithm~\ref{alg:low_level_topo}, lines 15-26).
\begin{algorithm}[t]
	\caption{Low-Level Topological Network Generation.}
	\label{alg:low_level_topo}
	\begin{algorithmic}[1] 
		\State \textbf{Input:} $M_{\text{g}} \gets \text{grid map}$
		\State \hspace{\algorithmicindent} $M^\prime_{\text{g}} \gets$ historical grid map
		\State \hspace{\algorithmicindent} $G^\prime_{\text{L}}(V^\prime_{\text{L}}, E^\prime_{\text{L}}) \gets$ historical low-level graph
		\State \hspace{\algorithmicindent} $isFirstMapping \gets$ exploration state
		\State \textbf{Output:} $G_{\text{L}}(V_{\text{L}}, E_{\text{L}}), clusters$
		\If{$isFirstMapping$ == True}
		\State $V_{\text{L}} = \textbf{uniformSampling}(M_g)$
		\State $E_{\text{L}} = \textbf{addEdge}(V_{\text{L}})$
		\State $isFirstMapping$ = False
		\Else 
		\State $M_{\text{diff}} = M_{\text{g}} - M^\prime_{\text{g}}$
		\State $V_{\text{L}} = \textbf{updateGraph}(V^\prime_{\text{L}},M_{\text{diff}}) $
		\State $E_{\text{L}} = \textbf{addEdge}(V_{\text{L}} - V^\prime_{\text{L}}, M_{\text{g}})$
		\EndIf
		\For{each $node$ in $V_{\text{L}}$}
		\If{$node.degree < 8$ }
		\If{$\textbf{checkFrontier}(node,M_{\text{g}}) == $True }
		\State $V_{\text{frontier}} \gets node$
		\EndIf
		\EndIf
		\EndFor
		\State $V_{\text{filterdFrontier}} = \textbf{filterFrontier}(V_{\text{frontier}})$
		\State $clusters = \textbf{clusterFrontier}(V_{\text{filteredFrontier}})$
		\For{each $cluster$ in $clusters$}
		\State $cluster.cPointNode = \textbf{centerPoint}(cluster)$
		\EndFor
	\end{algorithmic}
\end{algorithm}
\subsubsection{Topological Mapping}
Inspired by \cite{uniformSample}, we uniformly sample the occupancy grid map $M_g$ to construct the low-level topological graph (line 7). The sampling intervals are defined as $d_\text{sample} = k \cdot d_\text{grid}$, where $k$ is a user-defined integer multiple and $d_\text{grid}$ is the width of each grid cell in the occupancy grid map (the width of map resolution). Among the sampled grids, only those in the 'free' state can be used to generate nodes for the LTG, with each node capable of storing its own positional coordinates from the corresponding grid, as shown in \figref{fig_1}{a}.

Then, any two nodes \( n_1 \) and \( n_2 \) which are adjacent in the eight-directional neighborhood (up, down, left, right, and the four diagonal directions), with coordinates \((x_1, y_1)\) and \((x_2, y_2)\) on the occupancy grid map \( M_g \) are connected if all cells within a corridor of width slightly larger than the robot are free of obstacles. This corridor is centered on the interpolated line from \( n_1 \) to \( n_2 \), using linear interpolation parameters \(\Delta x = (x_2 - x_1)/k\) and \(\Delta y = (y_2 - y_1)/k\). The corridor's cells are defined by the set
\begin{equation}
	\begin{aligned}
		\left\{ \left( x_1 + i \cdot \Delta y + j \cdot \Delta x, y_1 + i \cdot \Delta x - j \cdot \Delta y \right) \right. \\
		\left. \mid i = -\left\lceil \frac{w}{2} \right\rceil, \ldots, \left\lceil \frac{w}{2} \right\rceil, j = 0, 1, \ldots, k \right\}
	\end{aligned}
\end{equation}
where \(i\) adjusts for the corridor's width \(w\), slightly larger than the robot's body width, and \(j\) indexes along the direct path between the nodes. Edges between nodes \(n_1\) and \(n_2\) are established only if all cells in the set are free of obstacles (line 8 and 13). This approach ensures that all edges in the graph represent paths that are actually navigable by the robot, excluding any inaccessible passages. (\figref{fig_1}{a}).

Finally, The LTG dynamically updates by comparing the current map $M_g$ with its previous version $M^\prime_{\text{g}}$, focusing updates on areas of change, denoted as $M_{\text{diff}} = M_{\text{g}} - M^\prime_{\text{g}}$ (line 11). After the initial construction of the LTG, subsequent updates are performed based on real-time changes in \(M_{\text{diff}}\), specifically targeting uniformly sampled cells within this differential area for topological mapping, rather than reassessing the entire map (line 12). This strategy optimizes computational resources and enhances the efficiency of the mapping process.

\subsubsection{Graph Frontier Structuring}

\begin{figure*}[t] 
	\centering
	\includegraphics[width=\textwidth]{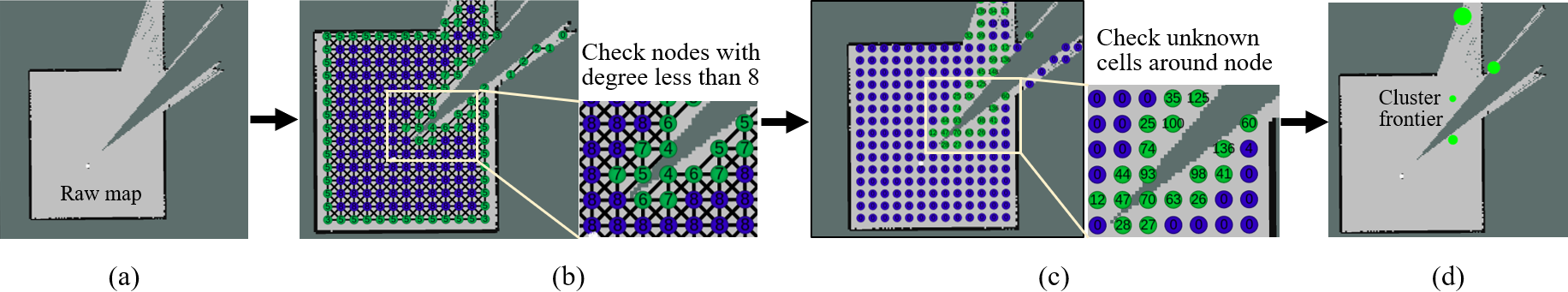} 
	\caption{Illustration of graph frontier structuring: (a) The occupancy grid map. (b)  Preliminary filtering of peripheral nodes with degrees less than 8 for frontier nodes extraction. (c) Diffusion from previously filtered nodes, expanding until reaching occupied cells or diffusion depth limit. (d) BFS-based clustering of frontier nodes for major area identification.}
	\label{fig_4}
\end{figure*}

In this module, as shown in Fig. \ref{fig_4}, we identify and cluster frontier nodes in the LTG. 
First, as shown in \figref{fig_4}{b}, we specifically target nodes with a degree less than eight for screening (line 16). These nodes are typically located at the graph's periphery adjacent to unknown or occupied areas, which indicates incomplete connectivity. By focusing on these nodes, we efficiently streamline the process for identifying frontier nodes, thereby enhancing the efficiency of the algorithm.

Subsequently, as shown in \figref{fig_4}{c}, we implement a diffusion-based approach starting from the grid cells of the previously identified peripheral nodes (line 17). Each node, represented by coordinates \((x, y)\) on the probability grid map, serves as the origin for diffusion. The diffusion process involves progressively spreading outward to a depth of \(d\), where \(d\) is a gradually increasing positive integer. The set of grid cells checked during this diffusion is defined by
\begin{equation}
	\{(x+i, y+j) \mid i \in [-d, d] \cap \mathbb{Z}, j \in [-d, d] \cap \mathbb{Z}\}
\end{equation}
The diffusion process for each node begins at a depth of \(d = 1\) and increases iteratively, where in each iteration, the algorithm evaluates a \((2d+1) \times (2d+1)\) grid of cells centered on the node's coordinates \((x, y)\). The number of unknown cells within this grid is tallied. Diffusion continues until one of two termination conditions is met: the detection of an 'occupied' cell within the grid, or the diffusion depth reaching the predefined maximum \(d_{\text{max}}\). Upon termination, the count of unknown cells at the current depth is recorded as the node's information value.

Finally, after determining the information values of all nodes on the LTG, the algorithm proceeds with filtering and clustering (lines 22-26). During the filtering step, nodes whose information values exceed a predefined threshold are designated as frontier nodes. After filtering, the Breadth First Search (BFS) algorithm is applied to the LTG to extract the connected components formed by the frontier nodes. These connected components serve as clusters of the frontiers. For each cluster, the node that is geographically closest to the centroid of all nodes within the cluster is selected as the candidate target node. This approach ensures that the selected targets are representative of the cluster's spatial characteristics and are optimally positioned for subsequent navigation tasks.

\subsection{The HTG Construction and Region Prioritization}

The process of the HTG module (the red block in Fig. \ref{fig_2}) is detailed in Algorithm~\ref{alg:high_level_topo}. This subsection comprises two main modules: the Graph Erosion Segmentation module (Algorithm~\ref{alg:high_level_topo}, lines 7-16) and the Trail Marker Mapping module (Algorithm~\ref{alg:high_level_topo}, lines 17-19).

\subsubsection{Graph Erosion Segmentation}

\begin{figure}[!t]
	\centering
	\includegraphics[width=\columnwidth]{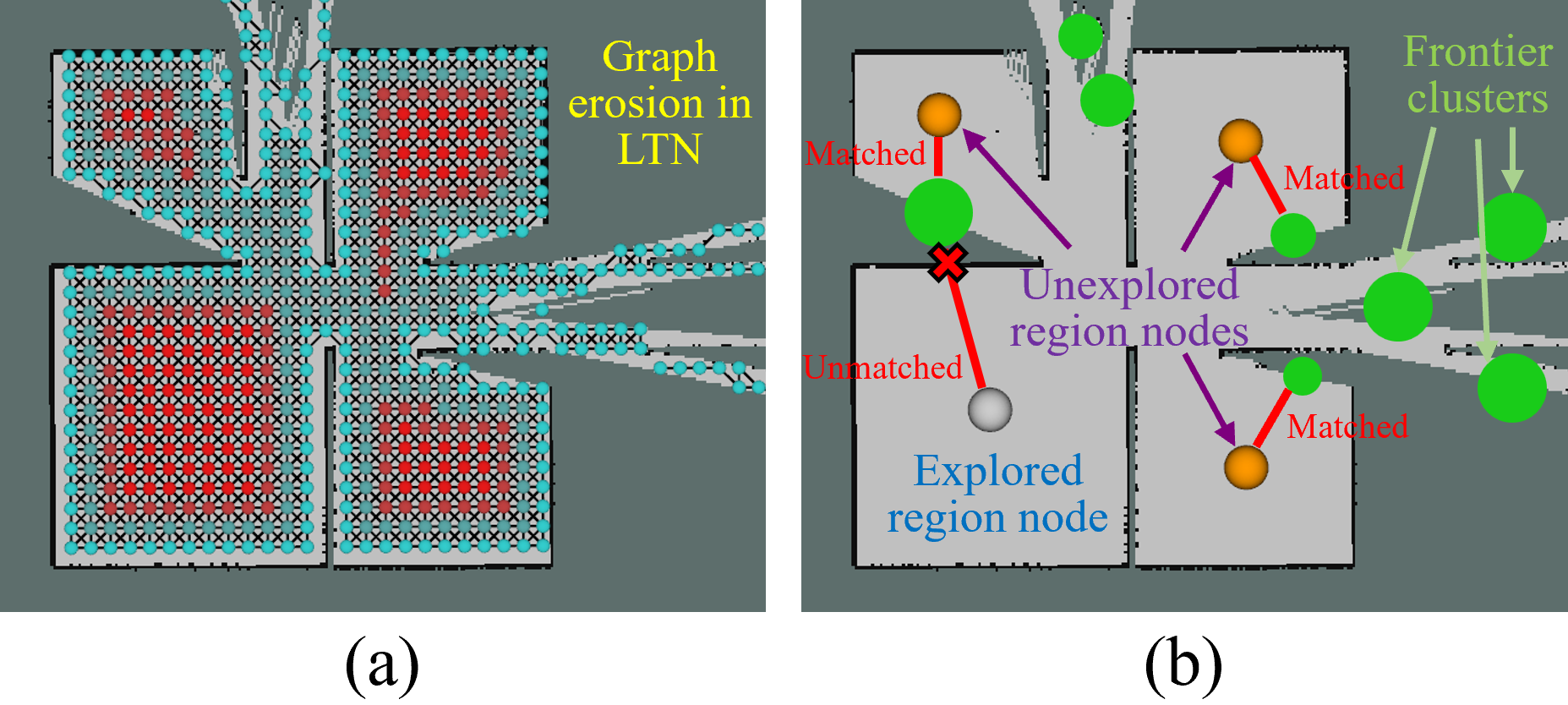}
	\caption{The process of graph erosion segmentation: (a) The erosion on LTG. The color gradient from lighter blue to darker red indicates the sequence of erosion, with lighter colors representing nodes eroded earlier. (b) The generation of region nodes. All region nodes, generated through the erosion process, are matched with frontier clusters to determine whether the region has been fully explored.}
	\label{fig_5}
\end{figure}

Inspired by computer graphics erosion techniques \cite{erosion}, we apply an erosion method to refine the LTG as shown in \figref{fig_5}{a}. First, the process starts by identifying initial subgraphs formed by removing nodes with degrees less than eight from the LTG. For each connected component \( C \) in these subgraphs, the erosion process is defined as

\begin{equation}
	C' = \text{ERODE}(C, \{v \in V(C) : \deg_C(v) < 8\})
\end{equation}

where \( C \) is a connected component of the graph, \( V(C) \) represents the nodes within \( C \), and \( \deg_C(v) \) is the degree of node \( v \) within component \( C \). The function \(\text{ERODE}(C, S)\) removes all nodes in the set \(S\) from component \(C\), along with their edges. 

\begin{algorithm}[t]
	\caption{High-Level Topological Network Generation.}
	\label{alg:high_level_topo}
	\begin{algorithmic}[1] 
		\State \textbf{Input:} $G_{\text{L}}, clusters \gets$ results of Algorithm 1
		\State \hspace{\algorithmicindent} $M_o \gets$ grid map
		\State \hspace{\algorithmicindent} $n \gets \text{number of erosion cycles}$
		\State \hspace{\algorithmicindent} $pose \gets \text{position of robot}$
		\State \hspace{\algorithmicindent} $V^\prime_{\text{trail}} \gets \text{historical trail marker nodes set}$
		
		\State \textbf{Output:} $G_{\text{H}}(V_{\text{H}}, E_{\text{H}})$
		
		\State $G_{\text{0}} = G_{\text{L}}$
		\For{$i$ = 1, 2, ..., $n$}
		\State $G_{\text{i}} = \textbf{graphErosion}(G_{\text{i-1}})$
		\EndFor
		\State $V_{\text{regions}} = \textbf{generateRegionNode}(G_{\text{n}})$
		
		\For{each $node$ in $V_{\text{regions}}$}
		\State $isUnexplored, clusterID =$
		\Statex \hspace{\algorithmicindent}$\textbf{checkCoverage}(node,M_o,clusters,d_{region})$
		\State $node.isUnexplored = isUnexplored$
		\State $node.clusterID = clusterID$
		\EndFor
		
		\State $V_{\text{trail}} = \textbf{generateTrailNode}(pose,G_{\text{L}},r_\text{trail},V^\prime_\text{trail})$
		\State $V_{\text{H}} = V_{\text{regions}} \cup V_{\text{trail}}$
		\State $E_{\text{H}} = \textbf{checkConnectivity}(V_{\text{H}},M_o,d_\text{edge})$

	\end{algorithmic}
\end{algorithm}

Subsequently, the erosion process is systematically applied to each connected component of the graph (lines 7-10). Erosion within a component ceases when no fully connected nodes (nodes with a degree of 8) are left, preventing excessive erosion that could lead to loss of important region structure information. A full cycle of erosion across all components constitutes one period. The process is repeated for the entire subgraph until no fully connected nodes remain, at which point the final subgraph is obtained. Each connected component of the final subgraph generates a region node for the HTG (line 11). The location of each region node is determined by the geometric center of all node positions within the connected component.

Finally, the exploration status of each region node is assessed by examining the connectivity with nearby frontier clusters within a predefined distance $d_{region}$. The Bresenham line algorithm \cite{Bresenham} is utilized to draw lines on the occupancy grid map from the grid cell corresponding to the region node's position to the centroid of each frontier cluster. As shown in \figref{fig_5}{b}, if all cells along the line are marked as free space in the occupancy grid, the connectivity is established, indicating that the convex polygonal region is still unexplored. If connectivity cannot be established with any frontier clusters, the region is considered explored.

\subsubsection{Trail Marker Mapping}

To enhance the navigational capabilities of the HTG, we introduce trail marker nodes that document the robot's trajectory during exploration (line 17). The trail marker nodes are generated at the robot's current position, with the generation condition based on two criteria: first, a minimum time interval must have passed since the last marker; second, there must be no existing trail marker nodes within a predefined radius \(r_{\text{trail}}\) of the current position. After creating a node, edges are created between the current trail marker node and any existing trail marker nodes within a predefined radius \(r_{\text{edge}}\) (\(r_{\text{edge}}\) $>$\ \(r_{\text{trail}}\)), where the connection is validated through Bresenham line-of-sight detection. 

Trail marker nodes are crucial for ensuring connectivity within the HTG. As show in \figref{fig_1}{b}, Region nodes represent the location and exploration status of convex polygonal region in map, while trail markers connect these regions to the robot (line 19). This connection allows for path cost calculations, even in incomplete maps. Without trail markers, the high-level network would consist of isolated region nodes, preventing efficient path planning and navigation.

\subsection{LAPF-Driven Motion Control within LTG}
The motion control module (the yellow block in Fig. \ref{fig_2}) for robot navigation includes two main components: Target Selection and Local Artificial Potential Field (LAPF).
\subsubsection{Target Selection}

\begin{figure}[!t]
	\centering
	\includegraphics[width=\columnwidth]{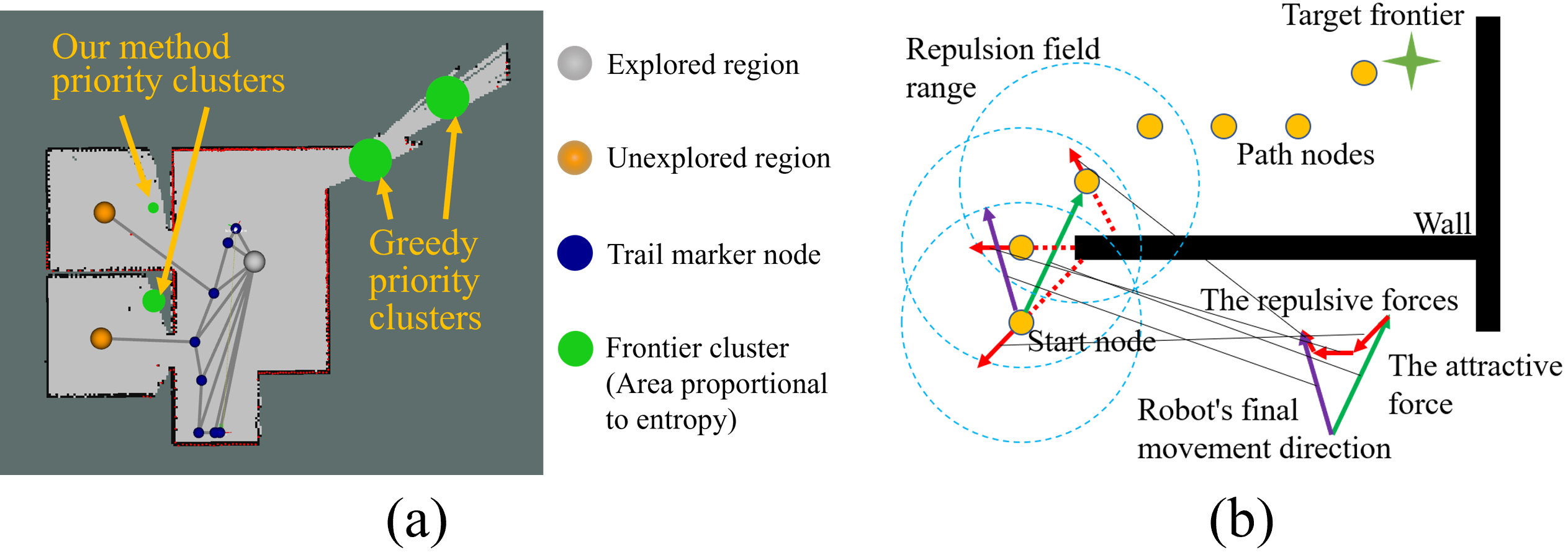}
	\caption{Illustration of target frontier cluster selection and LAPF: (a) Our approach focuses on exploring incompletely explored convex polygon regions, unlike traditional methods that prioritize high-entropy clusters, leading to reduced backtracking. (b) LAPF evaluates several initial nodes from the path nodes to integrate attraction and repulsion forces, determining the robot's optimal movement direction.}
	\label{fig_8}
\end{figure}

In the target frontier selection phase, as shown in \figref{fig_8}{a}, the algorithm checks for unexplored region nodes within the HTG. If at least one unexplored region node is present, the A* algorithm \cite{a_star} calculates path costs on the HTG, not based on Euclidean distance \cite{sample3,frontier17}, but on actual travel costs \cite{frontier15}, selecting the target as the frontier cluster matched with the unexplored region node that has the minimum path cost. Otherwise, the LTG is assessed for each frontier cluster's travel cost \( P \) and information gain \( I \). The total cost \( C \) for choosing a frontier is calculated as
\begin{equation}
	C = P - \alpha \cdot I
\end{equation}
where \(\alpha\) is a weighting factor. The frontier with the lowest \( C \) is selected as the target, optimizing the balance between exploration and computational efficiency.

\subsubsection{LAPF}

Upon selecting a target frontier, the nodes of path are computed using the A-star algorithm on the LTG. As shown in \figref{fig_8}{b}, the coordinates of the first three nodes on this path are denoted as \( p_0 \), \( p_1 \), and \( p_2 \). The unit vector $\vec{F}_{\text{attr}}$ of the attractive force is then calculated as
\begin{equation}
	\vec{F}_{\text{attr}} = \frac{p_2 - p_0}{\|p_2 - p_0\|}
\end{equation}

Repulsive forces are calculated for each node based on local occupancy data. The repulsion vector for node $i$, $\vec{F}_{\text{rep},i}$, is defined as the normalized sum of vectors pointing from the position $p_{\text{occ}}$ of occupied grid cell within a radius $d_{\text{sample}}$ to the node
\begin{equation}
	\vec{F}_{\text{rep},i} = \frac{\sum_{\text{occ}} (p_i - p_{\text{occ}})}{\|\sum_{\text{occ}} (p_i - p_{\text{occ}})\|}
	\label{eq:repulsion_vector}
\end{equation}

The robot's final movement direction, $\vec{D}_{\text{robot}}$, integrates these forces as
\begin{equation}
	\vec{D}_{\text{robot}} = w_1 \cdot \vec{F}_{\text{attr}} + \sum_{i=0}^{2} \left(\frac{w_2}{2^i} \cdot \vec{F}_{\text{rep},i}\right)
	\label{eq:motion_direction}
\end{equation}
where $w_1$ and $w_2$ are weights adjusting the influence of the attractive and repulsive forces, respectively. 

LAPF avoids local minima by using the third point of a dynamically updated path based on LTG, instead of using the final destination as the attraction point, which improves robot navigation efficiency over traditional potential field methods. Additionally, this direct motion control method enhances responsiveness and efficiency, particularly when navigating tight corners or frequently changing targets, providing significant improvements over the Dynamic Window Approach (DWA) \cite{dwa}, a widely adopted standard for local path planning and motion control in autonomous exploration that often suffers from slow response times and oscillations.

\section{Experimental Studies and Results}
\begin{figure}[t]
	\centering
	\includegraphics[width=\columnwidth]{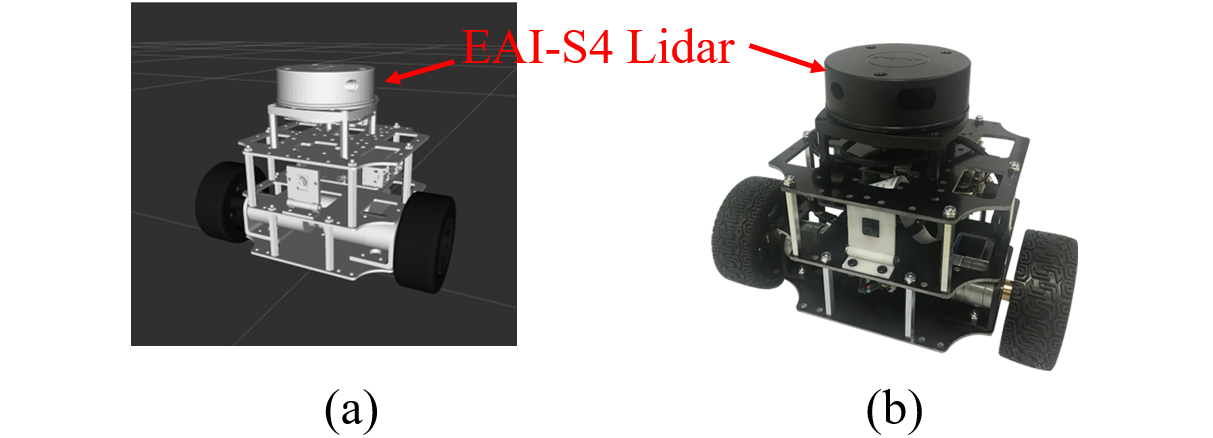}
	\caption{The robot platforms utilized in both simulation and real-world experiments, which operate with two-wheel differential drive control: (a) The simulation robot (b) The real robot.}
	\label{fig_6}
\end{figure}

Our experimental evaluation was conducted in both simulated environments and real-world settings. As shown in Fig.~\ref{fig_6}, the simulation utilized a robot model consistent with the real-world robot platform. Both setups, which used a two-wheeled differential robot equipped with the SAI-S4 360$^{\circ}$ LiDAR sensor, ensured that the simulation closely mirrored the real-world conditions. To assess the performance of our algorithm, we benchmarked it against three existing methods: RRT \cite{sample3}, E-LITE \cite{frontier17}, and 2D-SEG \cite{frontier15}.

\subsection{Simulation Experiments}

\begin{table}[t]
	\centering
	\begin{threeparttable}
		\renewcommand{\arraystretch}{1.0}  
		\caption{Parameters of experiments}
		\label{table_param}
		
		\begin{tabular}{@{}c|c|c@{}}
			\toprule
			Robot parameters & SLAM parameters & Motion controller \\
			\midrule
			Max speed: 0.25~m/s & Method: gmapping\cite{gmapping} & RRT: DWA \\	
			Max acc.: 2.5~m/s\(^2\)   & Resolution: 0.05~m & E-LITE: DWA \\	
			Max ang. speed: 1.0~rad/s & Update rate: 1~Hz & 2D-SEG: DWA \\		
			Max ang. acc.: 3.2~rad/s\(^2\) & Max scan range: 8.0~m & DART: LAPF \\
			\bottomrule
		\end{tabular}
\end{threeparttable}
\end{table}
Simulation experiments were performed in four standard structured environments. Metrics used included exploration time, traveled distance, and computation time, with experiment parameters standardized as shown in Table \ref{table_param}. Exploration terminated at 100\% coverage in simple environments (scene1 and scene2), and at 98\% in complex environments (scene3 and scene4). Each approach underwent 10 trials in each environment, starting from the same position, with detailed results in Table \ref{table_sim1}. These experiments were performed on a system equipped with an Intel Core i7-10510U CPU and 16GB of RAM.

\subsubsection{Exploration Performance}
As shown in Table \ref{table_sim1}, our method outperforms established benchmark methods, achieving the minimal exploration times and shortest movement distances in all trials.

\begin{figure*}[t] 
	\centering
	\includegraphics[width=\textwidth]{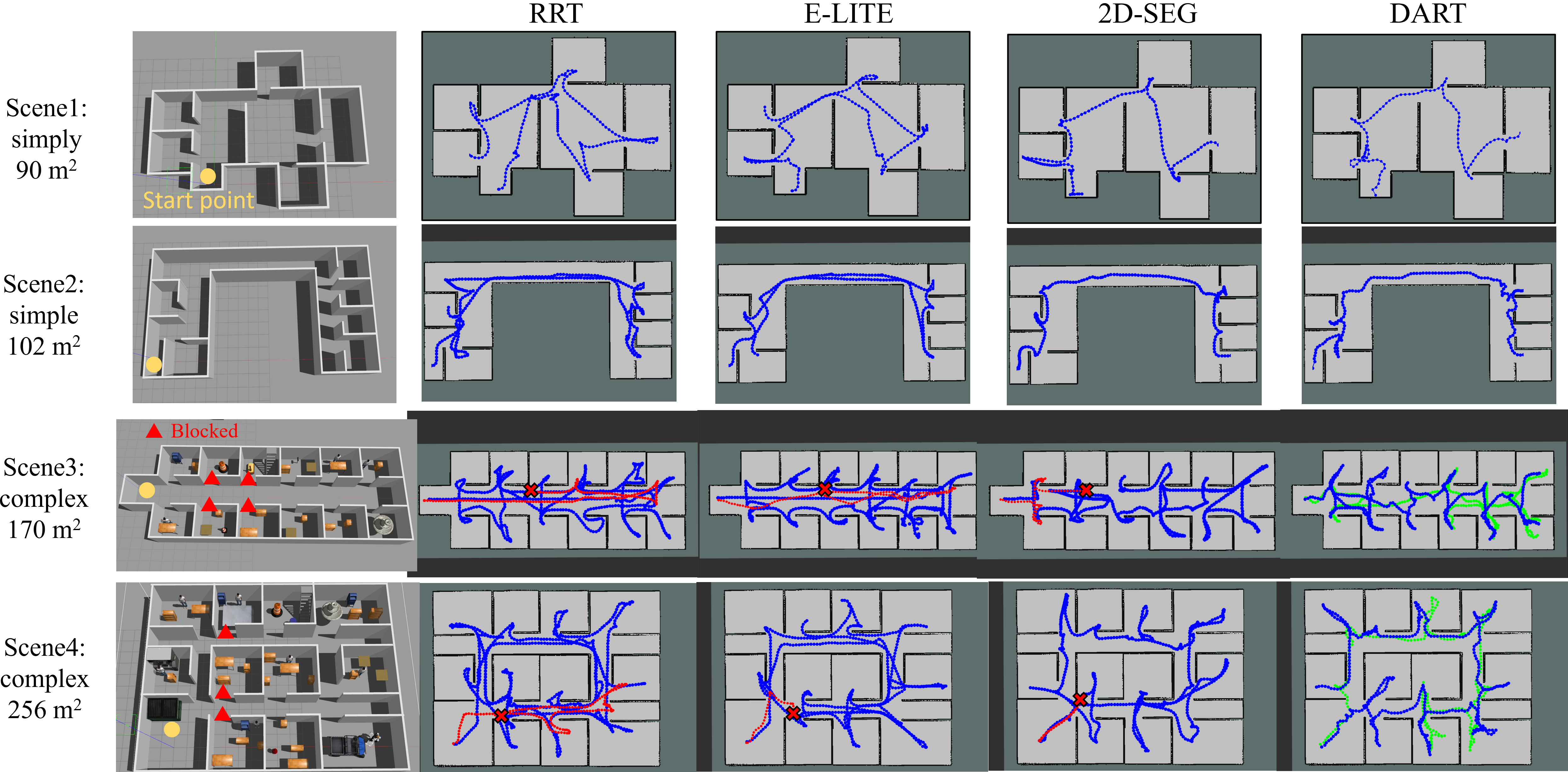} 
	\caption{Illustration of the trajectory lines for four algorithms across four scenarios, depicted as dot plots with a one-second interval between adjacent points. Higher point density indicates slower speeds, whereas lower density indicates faster movements. In the figure, blue trajectories represent paths in fully accessible environments, red trajectories denote failed planning in scenarios where doors are blocked but a gap allows detectable yet inaccessible region, and green trajectories illustrate successful path planning under the same blocked conditions.
	}
	\label{fig_11}
\end{figure*}

\begin{table}[t]
	
	\centering
	\begin{threeparttable}
		\renewcommand{\arraystretch}{1.0}  
		\caption{Results of Simulations in Four Environments}
		\label{table_sim1}
		\begin{tabular}{@{}c|c|ccc@{}}
			\toprule
			\multirow{2}{*}{Scene} & \multirow{2}{*}{Method} & \multicolumn{1}{c}{Exploration} & \multicolumn{1}{c}{Traveled} & \multicolumn{1}{c}{Computation} \\
			&  & \multicolumn{1}{c}{time (s)} & \multicolumn{1}{c}{distance (m)} & \multicolumn{1}{c}{time (ms)} \\
			\midrule
			& RRT & $211.7 \pm 12.7 $ & 43.2 $\pm$ 2.7 & 3877 $\pm$ 203 \\
			Scene 1 & E-LITE & $197.0 \pm 16.1$ & 40.6 $\pm$ 2.6 & 28.5 $\pm$ 1.0 \\
			90 m\(^2\) & 2D-SEG & $169.6 \pm 23.5$ & 36.1 $\pm$ \textbf{2.1} & 157.1 $\pm$ 1.4\\
			& DART & \textbf{128.8} $\pm$ \textbf{10.8}  & \textbf{30.9} $\pm$ 2.3 & \textbf{4.6} $\pm$ \textbf{0.2} \\
			\midrule
			& RRT & 317.4 $\pm$ 25.4 & 71.5 $\pm$ 3.5  &  4894 $\pm$ 207 \\
			Scene 2& E-LITE & 264.3 $\pm$ 17.5 & 67.3 $\pm$ 2.7 & 28.9 $\pm$ 1.9 \\
			102 m\(^2\)& 2D-SEG & 220.8 $\pm$ \textbf{17.1} & 42.3 $\pm$ 2.9 & 170.0 $\pm$ 2.8 \\
			& DART & \textbf{180.7} $\pm$ 19.6 & \textbf{41.2} $\pm$ \textbf{2.5} & \textbf{4.7} $\pm$ \textbf{0.7} \\
			\midrule
			& RRT & 669.8 $\pm$ 47.3 & 144.8 $\pm$ 20.3 & 5092 $\pm$ 205 \\
			Scene 3& E-LITE & 665.1 $\pm$ 40.2 & 131.1 $\pm$ 18.7 & 95 $\pm$ 6.3 \\
			170 m\(^2\)& 2D-SEG & 508.7 $\pm$ 25.2 & 100.1 $\pm$ 14.9 & 253.4 $\pm$ 24.2 \\
			& DART & \textbf{411.6} $\pm$ \textbf{23.6} & \textbf{95.1} $\pm$ \textbf{10.6} & \textbf{25.5} $\pm$ \textbf{2.4} \\
			\midrule
			& RRT & 844.6 $\pm$ 58.5& 181.6 $\pm$ 25.5 & 7405 $\pm$ 418 \\
			Scene 4& E-LITE & 830.7 $\pm$ 52.5& 164.1 $\pm$ 21.1 & 97.6 $\pm$ 9.7 \\
			256 m\(^2\)& 2D-SEG & 635.3 $\pm$ 41.7& 121.8 $\pm$ 16.5 & 212.8 $\pm$ 17.2 \\
			& DART & \textbf{411.3} $\pm$ \textbf{35.2} & \textbf{100.5} $\pm$ \textbf{13.2} & \textbf{28.9} $\pm$ \textbf{4.7} \\
			\bottomrule
		\end{tabular}

		\begin{tablenotes}
			\raggedright
			\setlength{\leftskip}{-0.4cm}
			\item {\fontsize{7pt}{7pt}\selectfont Data format: mean $\pm$ standard deviation (10 samples).}
		\end{tablenotes}
	\end{threeparttable}
\end{table}

\begin{figure}[!t]
	\centering
	\includegraphics[width=\columnwidth]{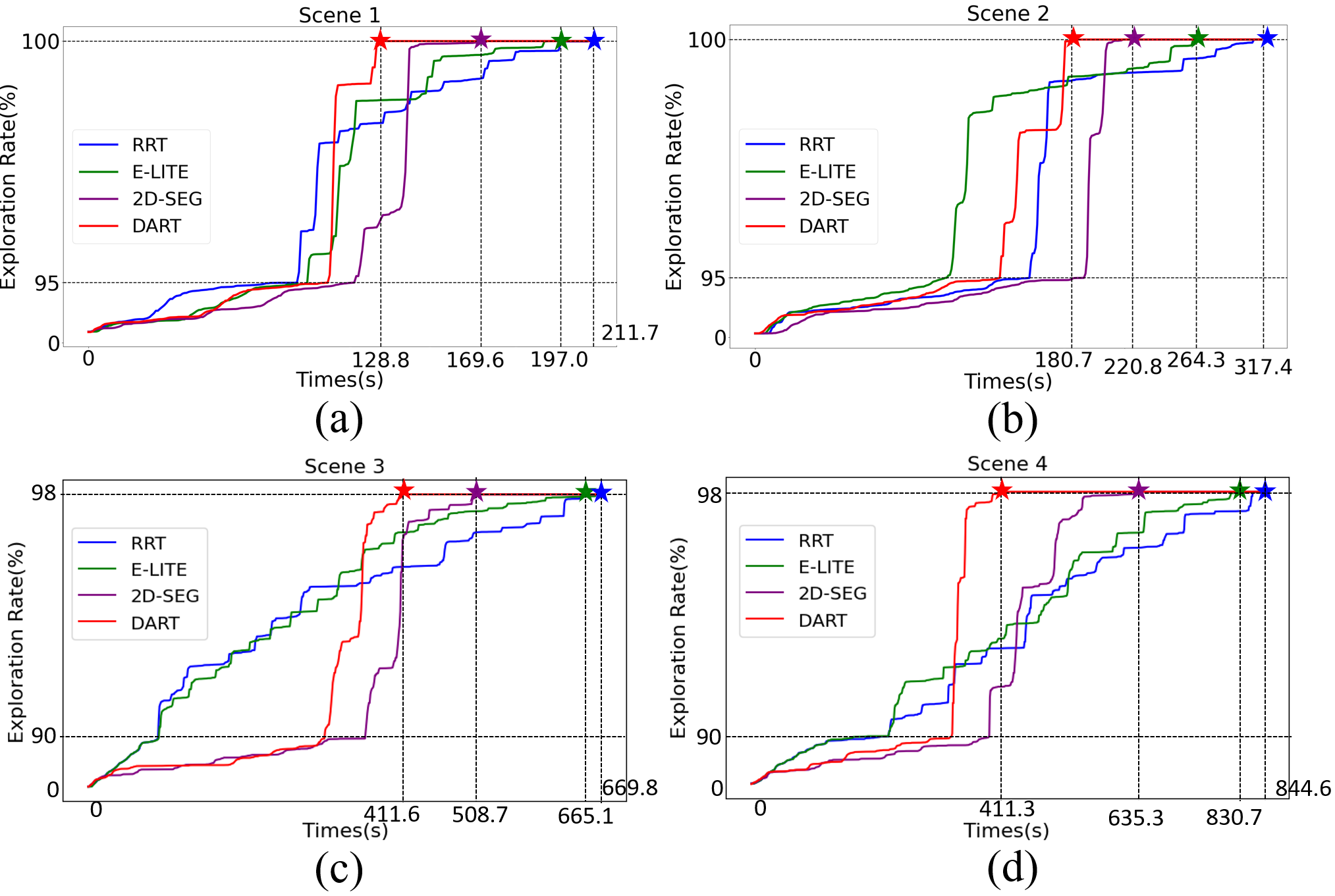}
	
	\caption{Results of exploration rates for all methods across four simulation environment: (a) Scene 1. (b) Scene 2. (c) Scene 3. (d) Scene 4.}
	\label{fig_10}
\end{figure}

As shown in Fig. \ref{fig_11} and Fig. \ref{fig_10}, the trajectories of different methods across various scenarios and the exploration rate over time are depicted. RRT and E-LITE, using a greedy strategy, show high initial exploration rates but incur backtracking due to revisiting missed areas, resulting in longer exploration times. Furthermore, when inaccessible frontiers are present, these methods often target them, leading to navigation failures. The 2D-SEG method segments the map to explore local frontiers first, performing better globally than RRT and E-LITE. However, in complex scenarios like scene 3 and scene 4, its segmentation is less effective, causing backtracking and failure when targeting inaccessible frontiers.

In contrast, our approach prioritizes the exploration of regions that have not been fully explored, significantly reducing the need for backtracking. This strategy ensures that the robot systematically completes each area before moving on to the next, thereby minimizing redundant movements and enhancing overall efficiency. Additionally, the LAPF in our framework increases the robot's average speed by dynamically adjusting its trajectory based on real-time environmental data, which facilitates rapid navigation through unexplored territories and efficient obstacle avoidance. Finally, the automatic exclusion of inaccessible frontiers, as discussed in Section IV.A through edge connectivity detection, ensures that the robot avoids dead ends, enabling successful exploration even in environments with inaccessible areas.

\begin{table}[t]
	\centering
	
	\makebox[\columnwidth]{
		\begin{threeparttable}
			\renewcommand{\arraystretch}{1.0}  
			\caption{Computational Time Distribution in Scene 4}
			\label{table_sim2}
			\begin{tabular}{@{}c|ccc|c@{}}
				\toprule
				\multirow{2}{*}{Method} & \multicolumn{1}{c}{Frontier check} & \multicolumn{1}{c}{Target selection} & \multicolumn{1}{c|}{Other time} & \multicolumn{1}{c}{Total} \\
				& \multicolumn{1}{c}{time (ms)} & \multicolumn{1}{c}{time (ms)} & \multicolumn{1}{c|}{cost (ms)} & \multicolumn{1}{c}{time (ms)} \\
				\midrule
				RRT & 5205 (70.3\%) & 2200 (29.7\%) & 0.0 (0.0\%) & 7405\\
				E-LITE & 97.1 (99.5\%) & 0.5 (0.5\%) & 0.0 (0.0\%) & 97.6\\
				2D-SEG & 19.7 (9.3\%) & 93.5 (43.9\%) & 99.6 (46.8\%) &212.8 \\
				DART & \textbf{14.2} (49.1\%) & \textbf{6.3} (21.8\%) & \textbf{8.4} (29.1\%) & \textbf{28.9} \\
				\bottomrule
			\end{tabular}
			
			\begin{tablenotes}
				\footnotesize
				\setlength{\leftskip}{-0.4cm}  
				\item {\fontsize{7pt}{7pt}\selectfont For the 2D-SEG method, "Other time cost" includes time spent on geometric segmentation of the map, while for our method, it includes the time required to construct the high-level Graph and run LAPF.}
			\end{tablenotes}
		\end{threeparttable}
		
	}
	
\end{table}

\subsubsection{Computational Efficiency}

As shown in Table \ref{table_sim1}, our method demonstrates superior computational efficiency due to its reliance on a uniformly down-sampled low-level topological Graph for exploration-related computations, achieving target point updates in milliseconds.

Table \ref{table_sim2} presents the comparative distribution of computational time across different exploration-related modules in Scene 4. The RRT method requires frequent tree expansions and information gain calculations for frontiers, increasing computational load. In contrast, E-LITE quickly selects targets using Euclidean distances and the size of boundary points, although it incurs high computation during frontier detection from raw map data. The 2D-SEG method utilizes innovative frontier extraction to reduce computation time and applies A-star path finding with path lengths as travel costs to enhance exploration efficiency, although this increases time costs. Additionally, its geometric segmentation, performed directly on the raw map, further adds to the computational load.

Our method enhances efficiency by rapidly identifying frontiers through a down-sampled topological Graph and calculating path lengths on this Graph, significantly speeding up the target selection process. Additionally, we employ an innovative graph erosion technique to rapidly construct a high-level Graph.

\subsection{Ablation experiments}

We conducted an ablation study to assess the individual contributions of our framework's components, detailed in Table \ref{table_abla}. Our system includes the low-level topological graph (LTG), essential for exploration tasks, the high-level topological graph (HTG), and the local artificial potential field (LAPF). Without HTG, the method assesses only the low-level graph frontiers for target selection. Without LAPF, the robot defaults to DWA for motion control.

\begin{table}[t]
	\centering
	\begin{threeparttable}
		\caption{Results of Ablation experiments}
		\label{table_abla}
		\renewcommand{\arraystretch}{1.0}
		\begin{tabular}{@{}c|ccc|ccc@{}}
			\toprule
			\multirow{2}{*}{Scene} & \multicolumn{1}{c|}{\multirow{1}{*}{Module}} & \multicolumn{1}{c}{Exploration} & \multicolumn{1}{c}{Traveled} & \multicolumn{1}{c}{Computation} \\
			& \multicolumn{1}{l|}{A\hspace{2.5mm}B\hspace{2.5mm}C} & \multicolumn{1}{c}{time (s)}  & \multicolumn{1}{c}{distance (m)} & \multicolumn{1}{c}{time (ms)} \\
			\midrule
			\multirow{4}{*}{Scene 3} 
			& \multicolumn{1}{l|}{\checkmark}           & \multicolumn{1}{c}{667.0 $\pm$ 45.6} & \multicolumn{1}{c}{137.2 $\pm$ 17.1} & \multicolumn{1}{c}{\textbf{17.8} $\pm$ \textbf{1.9}} \\
			& \multicolumn{1}{l|}{\checkmark\hspace{2.1mm}\checkmark} & \multicolumn{1}{c}{560.3 $\pm$ 37.9} & \multicolumn{1}{c}{107.1 $\pm$ 12.9}   & \multicolumn{1}{c}{25.0 $\pm$ 2.4} \\
			& \multicolumn{1}{l|}{\checkmark\hspace{6.3mm}\checkmark} & \multicolumn{1}{c}{555.1 $\pm$ 31.1} & \multicolumn{1}{c}{129.1 $\pm$ 15.1}   & \multicolumn{1}{c}{18.2 $\pm$ 2.0} \\
			& \multicolumn{1}{l|}{\checkmark\hspace{2.1mm}\checkmark\hspace{1.9mm}\checkmark} & \multicolumn{1}{c}{\textbf{411.6} $\pm$ \textbf{23.6}} & \multicolumn{1}{c}{\textbf{95.1} $\pm$ \textbf{10.6}}   & \multicolumn{1}{c}{25.5 $\pm$ 2.4} \\
			\midrule
			\multirow{4}{*}{Scene 4} 
			& \multicolumn{1}{l|}{\checkmark}                    & \multicolumn{1}{c}{820.0 $\pm$ 55.2} & \multicolumn{1}{c}{175.5 $\pm$ 23.1}  & \multicolumn{1}{c}{\textbf{20.2} $\pm$ \textbf{3.7}} \\
			& \multicolumn{1}{l|}{\checkmark\hspace{2.1mm}\checkmark} & \multicolumn{1}{c}{611.3 $\pm$ 45.9} & \multicolumn{1}{c}{115.3 $\pm$ 15.7}   & \multicolumn{1}{c}{27.7 $\pm$ 4.5} \\
			& \multicolumn{1}{l|}{\checkmark\hspace{6.3mm}\checkmark} & \multicolumn{1}{c}{649.0 $\pm$ 45.6} & \multicolumn{1}{c}{158.8 $\pm$ 20.5}  & \multicolumn{1}{c}{20.9 $\pm$ 3.9} \\
			& \multicolumn{1}{l|}{\checkmark\hspace{2.1mm}\checkmark\hspace{1.9mm}\checkmark} & \multicolumn{1}{c}{\textbf{411.3} $\pm$ \textbf{35.2}} & \multicolumn{1}{c}{\textbf{100.5} $\pm$ \textbf{13.2}}   & \multicolumn{1}{c}{28.9 $\pm$ 4.7} \\
			\bottomrule
		\end{tabular}
		\begin{tablenotes}[flushleft]
			\footnotesize
			\item {\fontsize{7pt}{7pt}\selectfont Data format: mean $\pm$ standard deviation (10 samples).}
			\item {\fontsize{7pt}{7pt}\selectfont A represents LTG, B represents HTG ,C represents LAPF.}
		\end{tablenotes}
	\end{threeparttable}
\end{table}

Ablation studies reveal that utilizing HTG reduces exploration redundancy, achieving a 34\% reduction in travel distance and a 25\% decrease in exploration time with a negligible increase in computational time. For the LAPF model, which relies on local node computations, the reduction in exploration time by 21\% is primarily attributed to higher average speeds compared to traditional motion control methods like DWA, with virtually no computational time cost. However, the travel distance sees only a modest 9\% decrease due to the lack of HTG analysis, leading to more redundant exploration. The minimal improvement in travel distance results mainly from reduced oscillations when the robot encounters corners or gets too close to walls. As a result, This highlights the importance of both comprehensive planning and fast-response motion control for optimizing exploration efficiency. 

\begin{figure}[!t]
	\centering
	\includegraphics[width=\columnwidth]{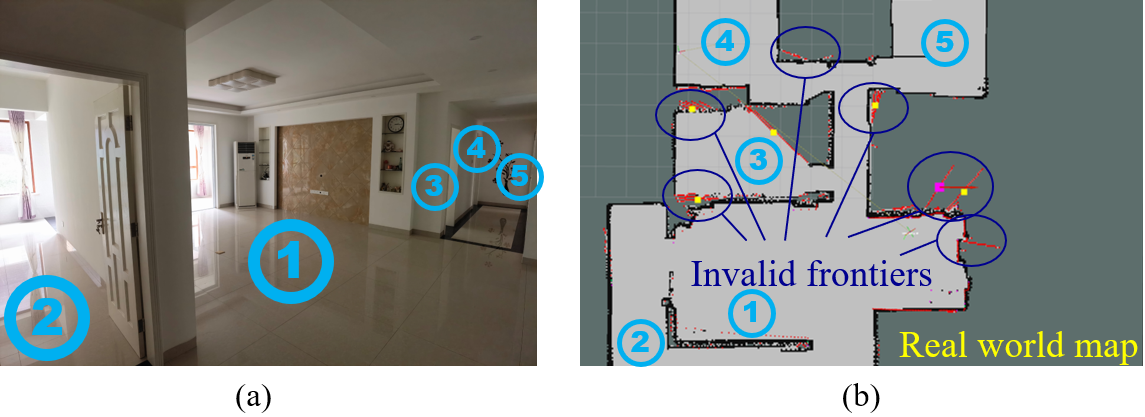}
	\vspace{-4mm}
	\caption{Illustration of experimental environment and SLAM mapping in a real-world environment, where numbers represent convex polygon region numbers: (a) Experimental environment with an area of 102 m\(^2\). (b) Mapping results in the real environment, showing a high presence of inaccessible, invalid frontiers.}
	\label{fig_9}
	\vspace{2mm}
\end{figure}

\begin{table}[t]
	
	\centering
	\makebox[\columnwidth]{
		\begin{threeparttable}
			\renewcommand{\arraystretch}{1.0}  
			\caption{The experimental results in real-world}
			\label{table_real}
			\begin{tabular}{@{}c | cccc@{}}
				\toprule
				\multirow{2}{*}{Method} & \multicolumn{1}{c}{Exploration} & \multicolumn{1}{c}{Traveled} & \multicolumn{1}{c}{Computation} & \multicolumn{1}{c}{Trial success} \\
				& \multicolumn{1}{c}{time (s)} & \multicolumn{1}{c}{distance (m)} & \multicolumn{1}{c}{time (ms)} & \multicolumn{1}{c}{rate (\%)} \\
				\midrule
				RRT & - & - & - & 0\% \\
				E-LITE & 240.8 $\pm$ 39.2 & 53.6 $\pm$ 16.1 & 37.7 $\pm$ 1.3  & 80\% \\
				2D-SEG & 202.5 $\pm$ 23.6 & 43.4 $\pm$ 7.5 & 189.3 $\pm$ 3.7 & 30\% \\
				DART & \textbf{151.7} $\pm$ \textbf{14.2} & \textbf{37.6} $\pm$ \textbf{2.6} & \textbf{5.1} $\pm$ \textbf{0.4} & 100\% \\
				\bottomrule
			\end{tabular}
			\begin{tablenotes}
				\footnotesize
				\setlength{\leftskip}{-0.4cm}  
				\item {\fontsize{7pt}{7pt}\selectfont Data format: mean $\pm$ standard deviation (10 samples).}
			\end{tablenotes}
			
		\end{threeparttable}
	}
\end{table}

\subsection{Real-word experiments}

To validate our method, we conducted real-world experiments, as shown in \figref{fig_9}{a}, with the robot shown in \figref{fig_6}{b}, equipped with the single-board computer Raspberry Pi 4B. All robot and SLAM parameters matched those in our simulations. Challenges in the real-world settings, shown in \figref{fig_9}{b}, included invalid frontiers caused by tire drift or reflections from glass, which posed risks of failure exploration.

As detailed in Table \ref{table_real}, the RRT model consistently failed as its historical random trees often extended beyond map boundaries, resulting in numerous detections of invalid frontiers when faced with map distortions. The E-LITE model, prioritizing frontiers with higher information gains, encountered fewer failures since invalid frontiers were less frequently identified as high-gain. The 2D-SEG model, by segmenting the map, also faced significant failure rates when invalid frontiers within the robot's current segment were clustered, misleading the exploration process. In contrast, Our method not only demonstrated superior exploration metrics but also highlighted the algorithm's robust adaptability to suboptimal map qualities. This adaptability is due to the algorithm's initial connectivity analysis within the low-level topological Graph, which excludes unreachable zones, ensuring reliable exploration outcomes even with mapping inaccuracies.

\section{Conclusion}

In this paper, we presented a dual-level topological framework for autonomous exploration in complex structured environments, combining the LTG and the HTG for strategic planning with our LAPF method. Our approach significantly enhanced exploration speed and computational efficiency. Initial simulation tests demonstrated reductions in exploration time and travel distance, along with improvements in computational efficiency. Following these simulations, ablation studies verified the critical contributions of each component, with real-world tests further demonstrating the robustness of our method. Future research could explore adapting this framework to more dynamic environments, such as those with moving obstacles or changing topologies. Overall, our method provides a robust and efficient solution for autonomous exploration, making it highly applicable to complex structured environments.


\bibliographystyle{IEEEtran}
\bibliography{references}\ 

\end{document}